\documentclass[letterpaper, 10 pt, conference]{ieeeconf}  

\IEEEoverridecommandlockouts                              

\usepackage{graphics} 
\usepackage{epsfig} 
\usepackage{mathptmx} 
\usepackage{times} 
\usepackage{amsmath} 
\usepackage{amssymb}  
\usepackage{graphicx}
\usepackage{color}
\usepackage{xspace}
\usepackage{subcaption}
\usepackage[T1]{fontenc}
\usepackage{hyperref}
\usepackage{multirow}
\usepackage{adjustbox}
\usepackage{authblk}
\usepackage{blindtext}
\usepackage{cite}
\usepackage{indentfirst}
\usepackage{afterpage}
\graphicspath{{./images/}}

\newcommand{\name}{CalTag\xspace}

\usepackage[explicit,noindentafter]{titlesec}
\titlespacing\section{0pt}{1.2ex}{0.5ex}
\titlespacing\subsection{0pt}{1ex}{0ex}

\title{\LARGE \bf
\name: Robust calibration of mmWave Radar and LiDAR using backscatter tags}

\author{Junyi Xu$^{1}$, Kshitiz Bansal$^{2}$, and Dinesh Bharadia$^{3}$}

\begin{document}

\affil{University of California San Diego}
\affil{\textit {jux006@ucsd.edu, ksbansal@ucsd.edu, dineshb@ucsd.edu}}

\maketitle
\thispagestyle{empty}
\pagestyle{empty}

\begin{abstract}

The rise of automation in robotics necessitates the use of high-quality perception systems, often through the use of multiple sensors. A crucial aspect of a successfully deployed multi-sensor system is the calibration with a known object typically named fiducial. In this work, we propose a novel fiducial system for millimeter wave radars, termed as CalTag. CalTag addresses the limitations of traditional corner reflector-based calibration methods in extremely cluttered environments. CalTag leverages millimeter wave backscatter technology to achieve more reliable calibration than corner reflectors, enhancing the overall performance of multi-sensor perception systems. We compare the performance in several real-world environments and show the improvement achieved by using CalTag as the radar fiducial over a corner reflector.

\end{abstract}

\section{Introduction}

Robotic automation is advancing rapidly, prompting the demand for high-quality perception systems. This evolution has made multi-modal sensing a standard in perception technology. Robots are typically equipped with sensors that offer complementary information, but they also introduce redundancy to a system essential for maintaining safety in life-critical applications, like autonomous driving. In a multi-modal sensing system, the most common sensors are LiDARs, cameras, and radars. Their data is integrated using sensor fusion techniques \cite{nabati2020radar, nabati2021centerfusion, ravindran2022camera, yao2023radar}.

Each sensor provides an independent view of the world in its coordinate system. To align data from different sensors, a transformation process, known as extrinsic calibration, is required. Extrinsic calibration uses the measurements from multiple sensors of a set of targets to compute a transformation matrix that maps each sensor's coordinate system to a common reference system. Extrinsic calibration methods are categorized into two types-- target-based methods \cite{domhof2019extrinsic,pervsic2017extrinsic,dhall2017lidar,yan2023joint} and targetless methods \cite{heng2020automatic, scholler2019targetless, cheng2023online}. Target-based methods use specific fiducial marker for each sensor to achieve high measurement precision. Whereas targetless methods rely on existing objects in the environment that are sensitive to all sensors. While targetless methods offer convenience and lower cost, they typically have lower calibration accuracy than target-based methods \cite{li2023joint}.

Incorporating multiple sensors benefits from selecting the most appropriate sensor for the encountered environment\cite{mohammed2020perception}. Due to its robustness against severe weather conditions and the efficiency in measuring object velocity \cite{nabati2021centerfusion, bansal2022radsegnet}, radar has become increasingly important in sensor fusion, particularly in the field of autonomous driving \cite{yao2023radar,bansal2020pointillism, bansal2023shenron} to achieve all-weather perception. This drives the demand for a more reliable and convenient extrinsic calibration strategy between the radar and other sensors. 

Calibration techniques with LiDAR and camera have been developed more extensively than with the radar, largely due to the simplicity of the fiducial marker principle and clear direction for amendment. Various LiDAR fiducial markers and associated techniques were proposed in \cite{bai2020lidar, yan2023joint}. The high sensitivity of LiDAR to sharp depth changes makes edges the primary choice for LiDAR fiducial designs. \cite{lee2020extrinsic} and \cite{yan2023joint} explore different edge shapes to enhance LiDAR calibration techniques. Camera fiducial designs often focus on corners, whose small sizes make them less affected by perspective and lens distortion. Built on top of this principle, the checkerboard method \cite{yan2023joint} and the circular fiducial markers \cite{daftry2013flexible} manipulate  with the shape of black and white regions to accommodate different conditions.

In contrast, research on fiducial design for radar calibration has been limited. Existing methods predominantly use the corner reflector as the fiducial marker \cite{jiang2023improving, lee2020extrinsic, pervsic2017extrinsic, domhof2019extrinsic, sugimoto2004obstacle}. The reliance on corner reflectors restricts advancements in accuracy, convenience, and effectiveness of radar calibration. Moreover, current radar calibration methods typically assume a clutter-free environment. Due to low radar signal resolution and lack of semantic information, clutters can heavily interfere with the fiducial's reflected signal, compromising the measurement accuracy or even leading to false detection. Solving this issue requires manual intervention to perform coarse calibration \cite{pervsic2017extrinsic}, but it does not guarantee success. These constraints limit the applicability and flexibility of radar calibration techniques. 

In this paper, we propose an altogether different radar fiducial design, \name, that uses millimeter-wave backscatter technology presented in \cite{bansal2022r} and optimize it to fit in radar calibration. \name completely removes the influence of clutters and background noise while maintaining the recognizable peak at the position of the fiducial. Through artificial frequency shift of the reflected wave, \name creates the effect that only a single object is detected in the predefined area, which makes it suitable for heavily cluttered environment and eliminates the need for coarse calibration.

We conduct extensive experiments that compare the performance of calibration using \name to using the corner reflector under different environments. The results prove that using \name can maintain a high calibration accuracy even in highly cluttered conditions that using a corner reflector fails. This observation opens new opportunities of deployment for radar systems. \name also provides more consistent calibration performance across different degrees of sensor rotation than the corner reflector. In either perspective, \name is more robust and reliable for radar calibration.

\section{Related Work}

Extrinsic calibration computes the necessary spatial transformation to align coordinate systems of multiple sensors such that they accurately represent the same positional information about an object. Fiducial design in target-based methods aims at maximizing the sensor's sensitivity to the marker. The corner reflector, known for its high Radar Cross-Section (RCS), compact size, and simple structure, is the most common radar fiducial \cite{jiang2023improving},\cite{lee2020extrinsic},\cite{pervsic2017extrinsic}. In open, low-clutter environments, the backscatter signal from a corner reflector is easily detectable, but multiple objects may be detected in cluttered environments. Missing semantic information, radar measurements alone cannot differentiate between the target and clutters.

Per{\v{s}}i{\'c} et al.\cite{pervsic2017extrinsic} proposed an initial coarse calibration method to address false detection. They first estimate the rough position of the corner reflector in the radar's coordinate system based on measurements from another sensor and the relative position between two sensors. Then, the correspondence between the radar and the LiDAR detection is accepted if only a single object is detected within the predefined area around the estimated position. This method is also used in \cite{lee2020extrinsic}. However, the manual measurements of relative sensor position are prone to error, especially when two sensors face different directions. Pairwise coarse calibration for all sensors is also impractical in large-scale radar calibration.

Besides coarse calibration, Sugimoto et al.\cite{sugimoto2004obstacle} adopted a dynamic approach by moving the corner reflector vertically across the radar plane several times to create a series of changes in intensity at the position of the corner reflector. However, this method requires a complicated mechanical design for the vertical movement.

Targetless calibration methods deploy more complicated techniques that can be classified into three categories \cite{yan2022opencalib}: edge registration, mutual information, and segmentation. The weak semantic information in the radar signal makes edge registration and mutual information difficult to apply to radar calibration. Izquierdo et al. \cite{izquierdo2018multi} leveraged segmentation-type radar calibration by using high-definition (HD) maps. This method selects static and high radar-sensitive structural objects as targets and estimates the transformation matrix using a DGNSS with real-time kinematic accuracy.

Due to the variability of environmental features, an increasing number of targetless calibration methods rely on the deep learning. Generally, the model incorporate both feature extraction and calibration into the end-to-end optimization. For instance, Sch{\"o}ller et al.\cite{scholler2019targetless} used two convolutional neural networks. The first network estimate the extrinsic matrix from the combined latent space information. The second network refines the estimation by minimizing residual errors of the first network. Cheng et al. \cite{cheng2023online} employed the YOLO model on radar feature extraction to combat low resolution and high noise in radar data. Deep learning-based targetless methods have apparent limitations in scalability and flexibility due to their fixed neural network structures and sensor configurations. Additionally, the performance of these models depends on the quality and variability of the training data, making it challenging to maintain high-quality feature detection in diverse and cluttered environments.
\section{Background}

\subsection{Radar Primer}
\textbf{Range estimation:} A frequency-modulated continuous wave (FMCW) radar uses frequency-modulated signals, often called chirp signals, to measure the round-trip time of reflected signals from the environment. Each chirp signal is a sinusoidal tone, whose frequency increases linearly with time. Within a chirp period, let's say $T$, the signal sweeps a fixed bandwidth $BW$. The reflected chirp signal from the environment is received by the radar and conjugate multiplied by the transmitted chirp, which results in a constant tone signal per reflection. The tone frequency is dependent on the round trip time of each reflection. The conjugated signal is sampled at $N$ samples per chirp, resulting in a fixed sampling rate of $f_s=\frac{N}{T}$.

Range information is directly embedded in the tone frequency. The distance $d$ between the Radar and the object can be calculated using the following equation
\begin{equation}
\label{radar_range_equation}
d=\frac{c \Delta f}{2\frac{BW}{T}}=\frac{ckf_s}{2\frac{BW}{T}}=\frac{ck\frac{N}{T}}{2\frac{BW}{T}}=\frac{ckN}{2BW},
\end{equation}
where $c$ is the speed of light, $\Delta f$ is the frequency of tone signal, and $k\in[0,1]$. We calculate $k$ by applying Fast Fourier Transform (FFT) over all samples in a chirp. $\frac{c}{2BW}$ is the range resolution, which determines the closest distance between two distinguishable objects by the radar.

\textbf{Angle estimation:} The angle of arrival $\theta$ is calculated using spatial Fourier Transform. The radar receiver consists of a linear antenna array. When a reflected wave reaches the radar's antenna array at a certain angle, it creates a linearly varying phase across the antenna array. With a carrier wavelength of $\lambda$ for the given radar, the distance between adjacent antennas is $\frac{\lambda}{2}$, so the path length difference between two adjacent antennas is $\frac{\lambda}{2}\sin(\theta)$. It gives a phase difference of $\sin(\theta)/2$. We can calculate the angle of arrival using $\theta=\sin^{-1}(2k)$, where $k$ is obtained by taking FFT over the readings from all antennas at one sample of signal and $k\in[0,1]$. This allows us to detect objects with a Field of View (FOV) from $-90^\circ$ to $90^\circ$. Please refer to \cite{dham2017programming} for more details on angle estimation using a linear antenna array.

\textbf{Doppler estimation:} When the object is moving relative to the radar at a velocity $v$, the Doppler effect creates a change in the frequency of the reflected wave that is proportional to the velocity of the object. This frequency change is calculated by $\frac{v}{c}f_c$, where $c$ is the speed of light and $f_c$ is the center frequency of the chirp. The frequency change is embodied as a phase change between consecutive chirps. We take FFT along multiple chirps to measure the frequency change encoded in the reflected wave and use it to calculate the relative velocity of the object.

\begin{figure*}[htbp]
    \centering
    \includegraphics[width=\linewidth]{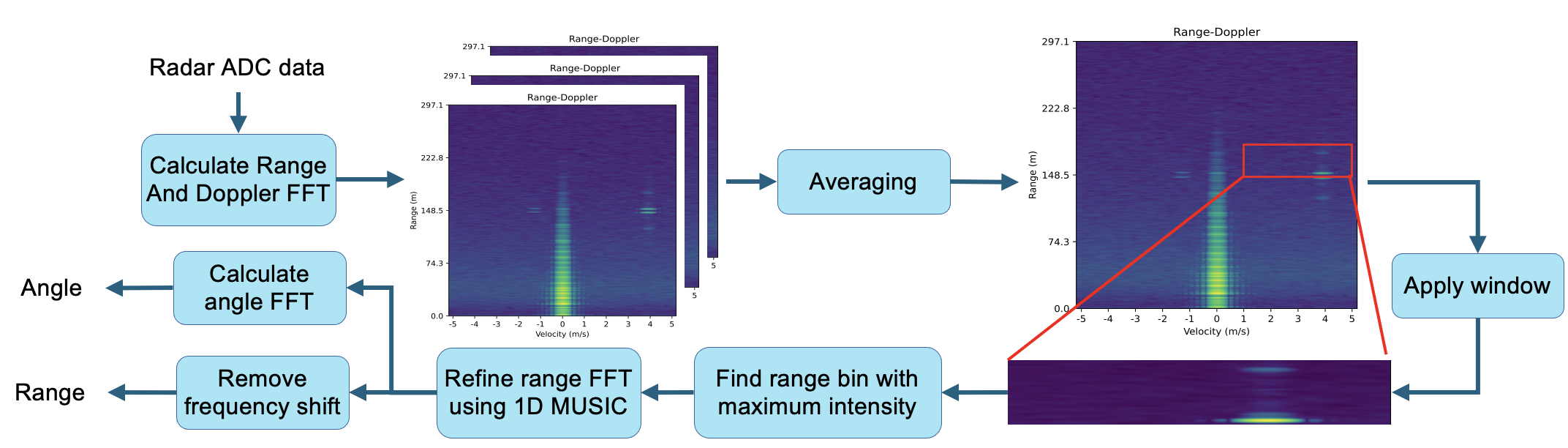}
    \caption{Flowchart for tag detection}
    \label{fig:flowchart}
\end{figure*}

\textbf{Range-Doppler FFT plot}: As shown in the lower part of Figure \ref{fig:harmonics}, the Range-Doppler FFT plot consists of Doppler FFT profiles at each range FFT bin. As an essential tool in our calibration method, it provides velocity information at different distances from the radar. The high intensity central line represents stationary objects at different distances. Peaks with positive Doppler frequencies represent objects approaching the radar. Peaks with negative Doppler frequencies represent objects moving away from the radar. \name leverages artificial Doppler generation to stand out from the heavy clutter at the zero Doppler region. Note that only the positive range axis is meaningful and participates in the further calculation, as shown in Figure \ref{fig:flowchart}.

\subsection{Backscatter Tag design principles}

\begin{figure}[htbp]
    \centering
    \includegraphics[width=\linewidth]{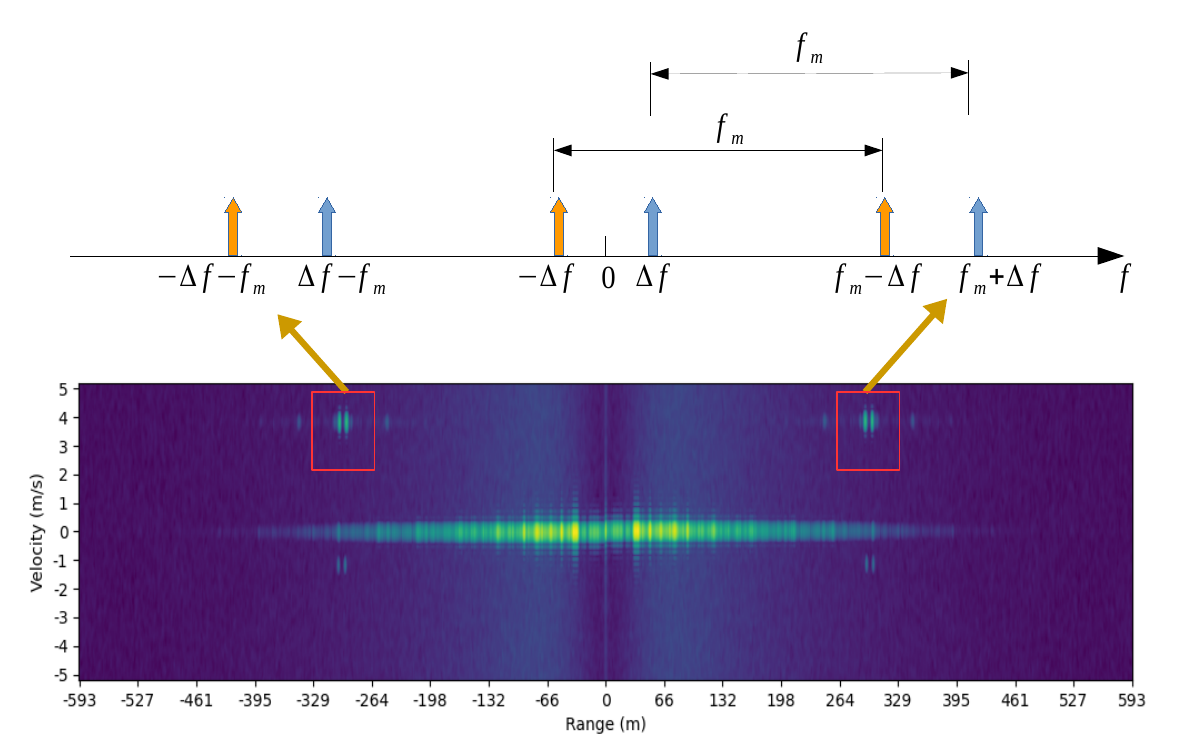}
    \caption{The upper plot shows the first harmonics. The lower plot shows the Range-Doppler FFT with modulated backscatter signals. High-intensity peaks inside red boxes are the modulated frequencies corresponding to harmonics at positive and negative frequency axes.}
    \label{fig:harmonics}
\end{figure}

The millimeter wave backscatter tags perform ON-OFF modulation with multiple cycles within a single chirp to create a certain frequency shift of the backscatter signal in the range domain\cite{bansal2022r}. When the \name's reflected signal appears at $\Delta f$, as shown in Figure \ref{fig:harmonics}, the in-chirp ON-OFF modulation creates harmonics at $\Delta f+n\times f_m$ using a modulation frequency of $f_m$. The first harmonics of the positive frequency component at $\Delta f$ appear at $\Delta f \pm f_m$, as indicated by blue arrows at two sides. Similarly, the negative frequency component at $-\Delta f$ creates first harmonics at $-\Delta f \pm f_m$, as indicated by orange arrows at two sides. The two high-intensity peaks at the positive range domain of the Range-Doppler FFT plot correspond to the two positive frequency components. By changing $f_m$, we can easily control the frequency of the first harmonics. Note that the Doppler shift of frequency can also be precisely controlled~\cite{vennam2023mmspoof}. When the on-off period of the modulation is not a multiple of the chirp duration, we will get a Doppler shift towards the non-zero Doppler regions. Therefore, we can shift the \name's signal to a static clutter-free region. 


\section{\name Design}

In this section, we will go over how \name performs LiDAR and radar calibration. 

\subsection{Detection}
The procedures for detecting the range and the angle of arrival associated with the \name are shown in Figure \ref{fig:flowchart}. We detect the frequency bin with the maximum power on the range-doppler FFT plot within a predefined region. Given that there is a harmonic at $\Delta f+f_m$, we set the predefined frequency range to be $[f_m, f_m+{\Delta f}_{max}+\alpha)]$, where ${\Delta f}_{max}$ is the frequency associated with the maximum distance from the radar that we place the \name and $\alpha$ is a constant frequency offset for the tolerance of detection error. We also set the predefined frequency range in the velocity domain from $\beta$ to maximum Doppler frequency, where $\beta$ is a customized frequency that prevents the region from including the zero frequency line. The red rectangular box in Figure \ref{fig:flowchart} represents the Range-Doppler FFT section for detecting the shifted frequency. A zoomed-in version of the segmented region is presented in the flowchart as well. As expected, within the segmented area, the bin with the highest power corresponds to the shifted frequency component.

Subsequently, we take the Fourier transform along the samples from multiple antennas at the detected range frequency bin to obtain the angle FFT profile. The angle frequency bin with the highest power gives the angle of arrival of the backscatter signal. The original range frequency bin, which reflects the \name's actual range, is calculated by subtracting the modulation frequency from the detected frequency. Since both the positive and negative frequency components associated with the backscatter signal are shifted by $f_m$, as shown in Figure \ref{fig:harmonics}. An alternative way of calculating the actual range is by detecting both peaks at two sides of the modulation frequency and compute the distance corresponding to half of the frequency difference:
$$
\frac{(f_m+\Delta f)-(f_m-\Delta f)}{2} = \frac{\Delta f-(-\Delta f)}{2} = \Delta f. 
$$

\subsection{Improving range accuracy through super resolution}

Calibration with \name needs a high enough radar sampling frequency to create a clutter-free frequency range. Meanwhile, we need to estimate the range with millimeter level precision. In the Fourier transform, the FFT bin width represents the frequency resolution and is determined by the sampling frequency $f_s$ and the number of FFT points. 

However, using a very large number of FFT points requires extensive computational resources. To tackle this issue, we use a combination of FFT and super-resolution method MUltiple SIgnal Classification (MUSIC). Since the shifted frequencies will fall inside a known range of frequencies. Instead of searching from $-\frac{f_s}{2}$ to $+\frac{f_s}{2}$, we use MUSIC to search only within the frequency range that the shifted frequencies may occur. Due to the narrower frequency spectrum to focus on, this method also allows us to search frequencies using a finer resolution. The 2D FFT with small number of FFT points  is first used to determine the rough Range-Doppler frequency bin where the shifted peaks appear. Then, a 1D MUSIC is applied to the selected Doppler bin to search frequencies only in the predefined range of frequencies along the range axis with a much higher resolution. Finally, the angle of arrival is calculated by applying 1D MUSIC over samples of multiple antennas at the identified range.

\subsection{Fiducial detection using LiDAR}
We set the LiDAR as the reference sensor and look for the transformation matrix of the radar's point cloud to best match the LiDAR's point cloud. Since Radar gives range and angle information $(r,\theta)$ in 2D, we discard the height information in LiDAR's point cloud, projecting the 3D point cloud into a 2D Bird Eye View (BEV). Since the fiducial appears as a dense set of points in the BEV when the fiducial is put in an empty space, we use Density-based spatial clustering of applications with noise (DBSCAN) to capture the fiducial by finding the cluster with the smallest standard deviation in the $x$ and $y$ direction. The center of the cluster is a close estimate of the center position of the fiducial.

\subsection{Calibration Procedure}
\name is placed at multiple locations to obtain point clouds in the radar and LiDAR coordinate systems respectively. An example is given in Figure \ref{fig:calib_comparison}.

The extrinsic matrix consists of a $2\times 2$ rotation matrix $R$ and a $2 \times 1$ translation matrix $t$.
$$
T=
\begin{bmatrix}
&R &t \\
&0 &1 \\
\end{bmatrix}
$$

The known correspondence between points in the source point cloud (radar) and points in the reference point cloud (LiDAR) allows us to use the Kabsch algorithm~\cite{kabsch1976solution} for calibration. 
We first translate the centroid of the radar point cloud matrix $\mu_R$ and LiDAR point cloud matrix $\mu_L$ to the origin in their respective coordinate system and recalculate translated point clouds, 
\begin{align*}
\mu_R &= \frac{1}{N}\sum_{i=0}^{N}q_i & \mu_L &= \frac{1}{N}\sum_{j=0}^{N}p_j \\
q'_i &= q_i - \mu_R & q'_j &= p_j - \mu_L
\end{align*}
where $q_i$ is the position of $i$th point in the radar point cloud, and $p_j$ is the position of the $j$th point in the LiDAR point cloud, $N$ is the total number of valid \name position. $q'$ and $p'$ denote the translated radar and LiDAR point clouds.

The cross-covariance matrix is calculated from the pairwise product of the translated point clouds and then decomposed using the Singular Vector Decomposition (SVD):
$$W = \sum_{(i,j)\in N}{q'}_i^Tp'_j = UDV^T.$$
The rotational matrix is computed as $R=UV$. The translation matrix is computed as $t = \mu_L - R\mu_R$.

\section{Implementation}

\subsection{Experiment setup}
\begin{figure}[h]
\hspace{0.1\linewidth}
\begin{subfigure}[b]{0.3\linewidth}
    \centering
    \includegraphics[width=\linewidth]{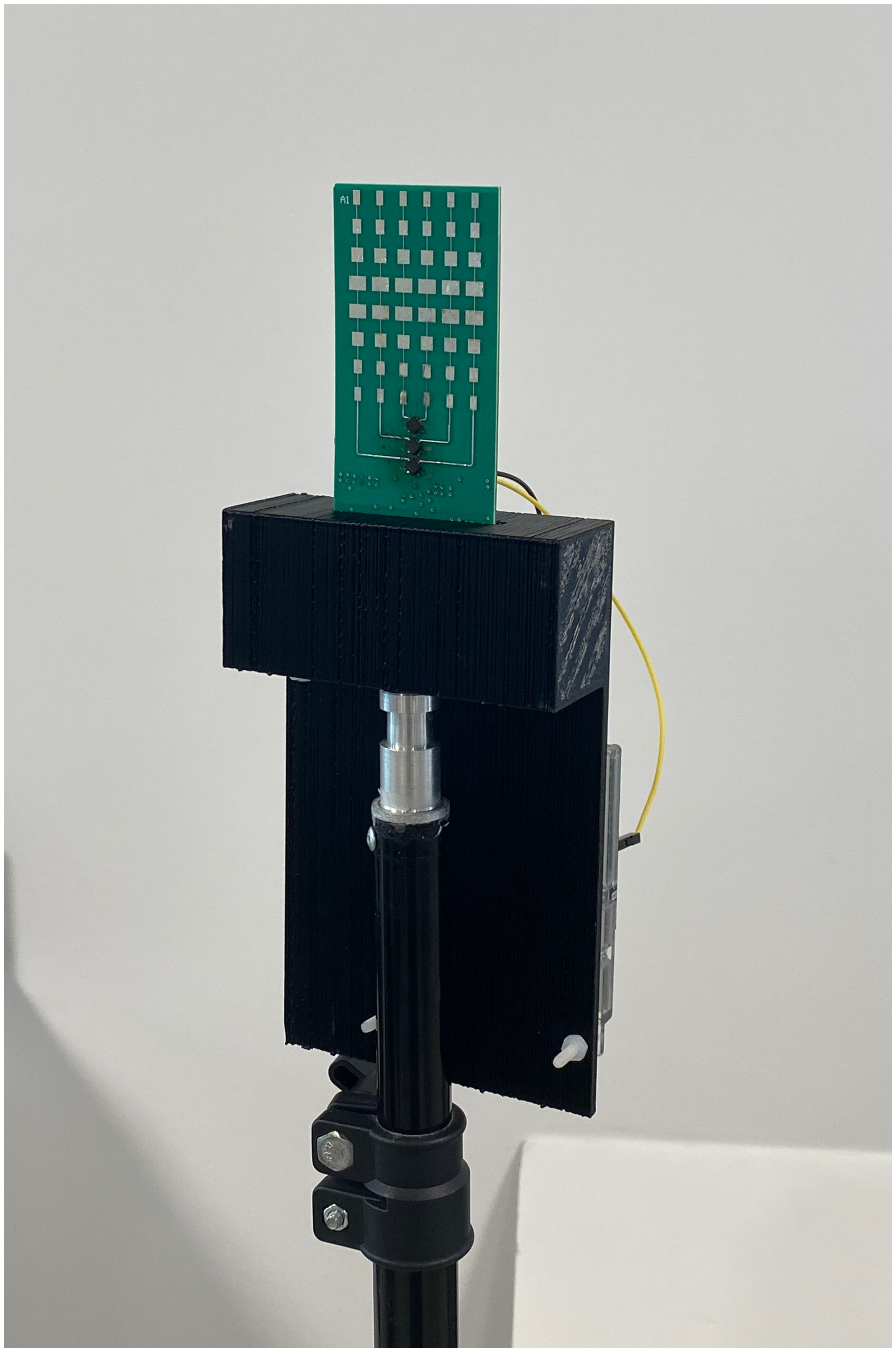}
    \subcaption{\name on a 3D printed base}
    \label{fig:mounting_top}
\end{subfigure}
\hspace{0.1\linewidth}
\begin{subfigure}[b]{0.34\linewidth}
    \centering
    \includegraphics[width=\linewidth]{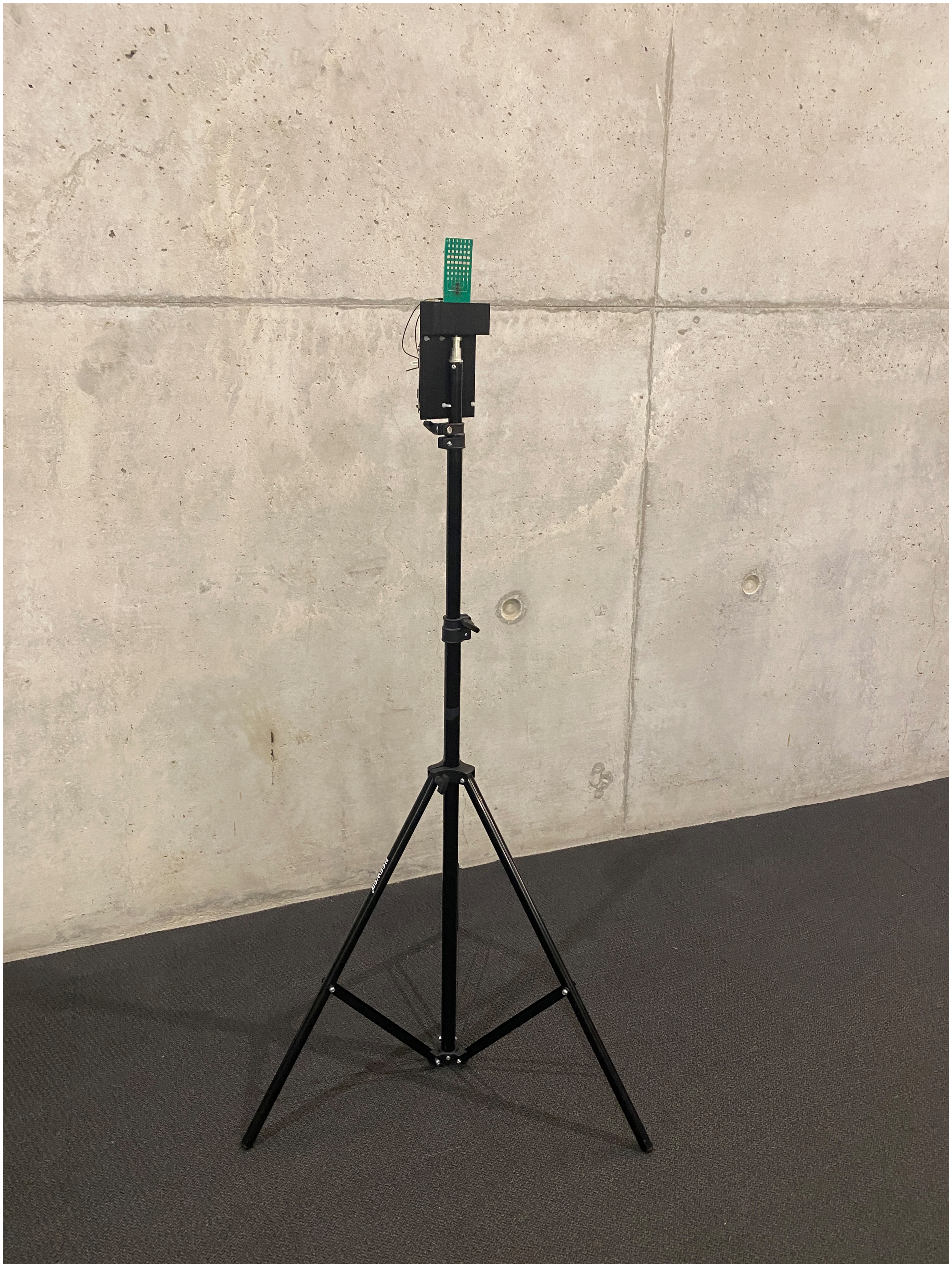}
    \subcaption{Complete fiducial structure}
    \label{fig:mounting_whole}
\end{subfigure}
\hfill
\caption{\name mounting structure}
\label{fig:mounting}
\end{figure}

\begin{figure*}[t]
    \centering
    \includegraphics[width=0.9\textwidth]{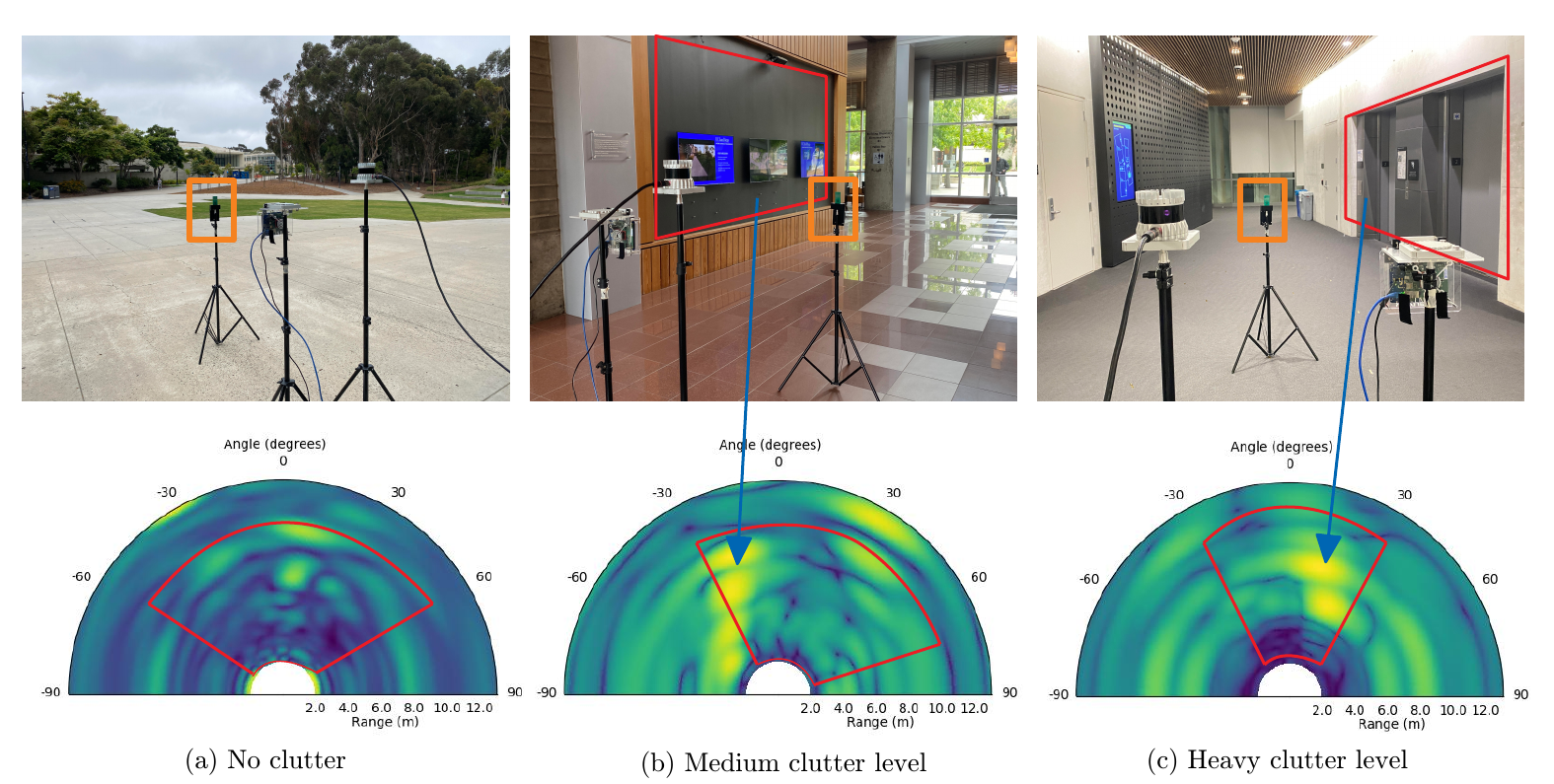}
    \caption{Comparison of three clutter levels. The top images show the environments, in which regions within the red polygons are clutter sources. The orange box marks the position of the \name. Bottom radar plots show the clutter sources and other background noise in polar coordinates. The red fan-shaped boundary marks the data collection region. The blue arrow maps the position of the clutter source from the image to the radar plot.}
    \label{fig:clutter_comparison}
\end{figure*}

In our experiments, we use an INRAS 24-GHz FMCW Radar \cite{radarbook2}, an OS1 Ouster LiDAR \cite{os1lidar}, a corner reflector, and a \name. The \name is placed on top of a 3D printed base and mounted on a tripod, as shown in Figure \ref{fig:mounting_top} and \ref{fig:mounting_whole}. The whole structure is kept in front of the LiDAR-Radar system, as shown in Figure \ref{fig:clutter_comparison}. We manually adjust the height of the fiducial to align with the height of the Radar. 

\subsection{Radar parameters}
We operate the radar with 250MHz bandwidth, achieving a range resolution of 0.6m. Since the tag's modulation frequency is set to 500KHz, we use 2MHz sampling frequency, which is four times the modulation frequency, to position the shifted peaks around half of the maximum detectable range. This is achieved by using 992 samples per chirp and a chirp duration of 496ms. With 250 MHz bandwidth and 992 samples per chirp, the Radar's maximum detectable range is 297.6m. This means the \name's distance is shifted by 148.8m, effectively isolated from potential clutters.

\subsection{Radar signal processing parameters}
For the rough peak detection, we use 1024-point 2D FFT along the range and the Doppler axis to locate the shifted frequency. Subsequently, in the refined range measurement, we use 1D MUSIC with 2cm resolution along the range axis at the identified Doppler frequency of the peaks. 1D MUSIC with $0.25^\circ$ degree resolution along the angle axis is then used to measure the angle of arrival at the identified range.

\subsection{Calibration error measurements}
The transformation error is measured through the root mean square error of the transformation outcome for all fiducial positions as follows,
$$
\epsilon=\sum_{i=1}^{N}||p(y_k^L)-Tg(y_k^R)||^2,
$$
where N is the total number of tag position, $g(y_k^R)$ converts the detected polar coordinates of the fiducial $(r_k,\theta_k)$ to 2D Euclidean coordinates $(x,y)$ and $p(y_k^L)$ projects the 3D LiDAR point cloud into point cloud in 2D BEV.

\section{Evaluation}

\begin{figure*}[t]
    \centering
    \includegraphics[width=\textwidth]{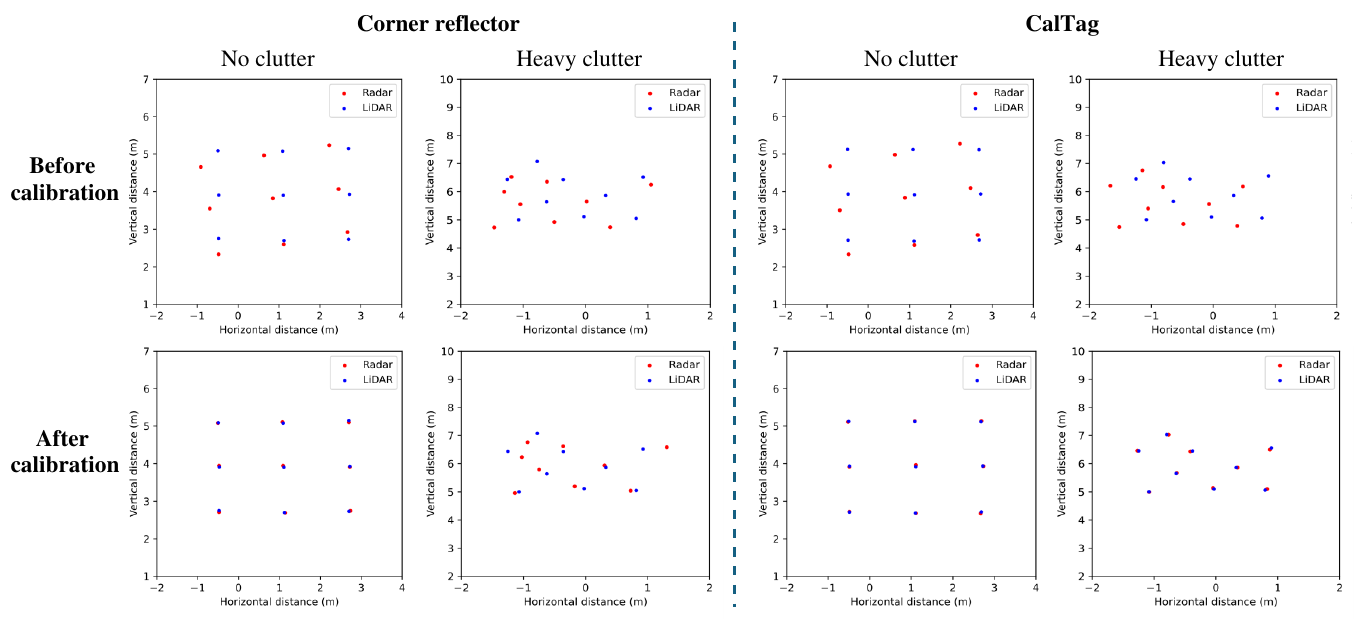}
    \caption{Comparison of the calibration results between using corner reflector and using \name in an environment with and without strong clutters. Red points are radar point clouds. Blue points are LiDAR point clouds.}
    \label{fig:calib_comparison}
\end{figure*}

In this section, we will present a comprehensive evaluation of our proposed calibration with \name. We conduct experiments in different levels of cluttered environments and with different relative positions of Radar and LiDAR. In each scenario, we compare the results from the \name with those from a corner reflector. Since LiDAR serves as the reference sensor, we treat the tag positions detected by LiDAR as the ground truth. For each experiment, data is collected with multiple fiducial positions and divided into training and testing sets. Calibration is performed on the training set and then evaluated on the testing set.

\textbf{Corner reflector baseline:} For the detection of corner reflectors, we followed the method in \cite{pervsic2017extrinsic} to perform coarse calibration with a window size of 5$^\circ$ in angle and 10cm in range and selected the peak in the window as the target. 

\textbf{Experiment procedure:} We collect 9 valid positions for each fiducial (correctly detected by both sensors). Among the 9 positions, we randomly select 6 positions to calculate the extrinsic matrix that matches the radar point cloud to the LiDAR point cloud and assess the calibration accuracy on the remaining 3 positions. This process is repeated 50 times.

\subsection{Calibration under different clutter levels}
Robotic environments with LiDAR-Radar setups can be highly complex and cluttered. To evaluate our method's reliability, we conducted calibration experiments across three clutter levels, comparing the performance of using \name to using a corner reflector. Given that angle measurement errors in coarse calibration heavily impact calibration accuracy with the corner reflector, we kept the radar and LiDAR face roughly the same direction throughout the experiment.

\textbf{Description of clutter levels:} Figure \ref{fig:clutter_comparison} illustrates three clutter levels using both real-world images and radar range-angle FFT plots. The red fan-shaped boundary marks the region in which we put the fiducial and collect data, namely the region of interest. Figure \ref{fig:clutter_comparison}a shows the outdoor case where no prominent source of clutter is close to the data collection area. The bottom radar plot demonstrates that only weak environmental noise exists. Figure \ref{fig:clutter_comparison}b shows an indoor environment with open space. There are some clutter sources near the edge of the data collection area, such as the wall to the left of the fiducial. The wall appears in the radar plot as the vertical high-intensity section at the left boundary of the region of interest, so its impact is insufficient to affect fiducials placed in the middle of the region. Figure \ref{fig:clutter_comparison}c shows another indoor case with strong sources of clutter. The metallic elevator gates to the right of the fiducial reflect much more electromagnetic waves, resulting in a larger impact. In the bottom radar plot, we observe both peaks corresponding to two clutter sources and strong side lobes.

\begin{figure}
    \centering
    \includegraphics[width=0.9\columnwidth]{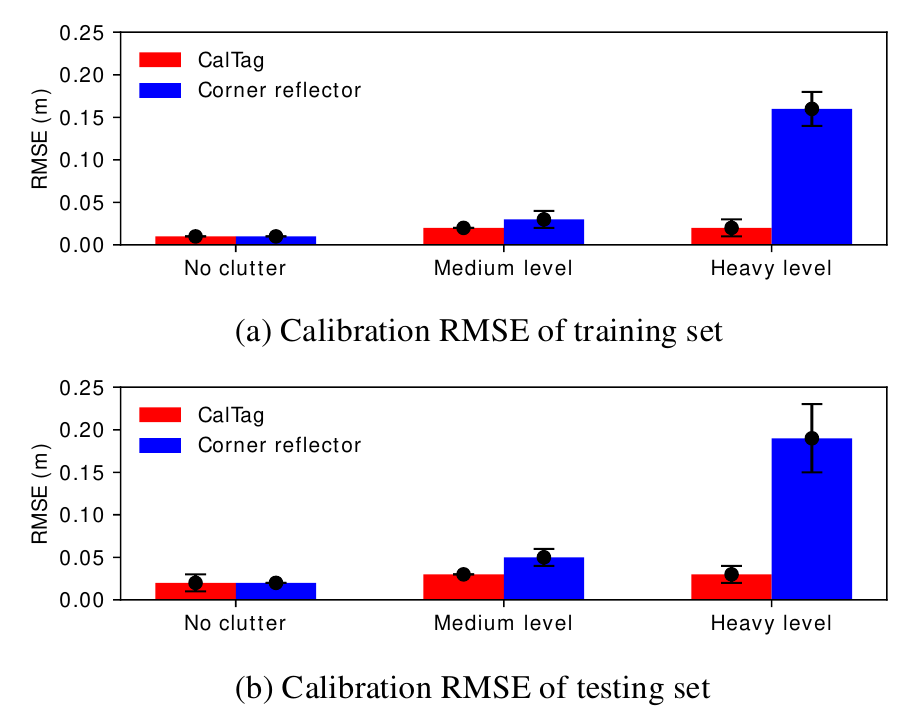}
    \caption{Training and testing calibration error with \name and corner reflector as the fiducial under different clutter levels}
    \label{fig:calib_rmse}
\end{figure}

\textbf{Qualitative result:} Figure~\ref{fig:calib_comparison} displays calibration outputs using two fiducials across different clutter levels. The pre-calibration point clouds represent fiducial positions in radar and LiDAR coordinates. The post-calibration point clouds show radar data transformed into the LiDAR coordinate system using the estimated extrinsic matrix. The extent of overlapping between two post-calibration point clouds indicates the calibration performance. There is little difference between the performance under low clutter level. However, strong clutters have a huge impact on the performance when the corner reflector is used but still have little influence on the performance when the \name is used.

\textbf{Quantitative result:} Figure \ref{fig:calib_rmse} compares the RMSE while using \name as the fiducial to the RMSE while using the corner reflector as the fiducial, for both training and testing sets. With \name, calibration RMSE is below 0.03m for the training set and below 0.05m for the testing set across all clutter levels. This demonstrates the robustness of \name against clutters. As expected, the RMSE of calibration using the corner reflector is highly dependent on the clutter level. It reaches the same accuracy as using \name only in clutter-free conditions. As the clutter level increases, the RMSE rises, but the performance is still acceptable when the clutter has limited influence on the region of interest. However, calibration using the corner reflector completely fails in a heavily cluttered environment. The RMSE shoots beyond 0.15m for both training and testing sets. Huge variation in RMSE conveys unreliability of using a corner reflector as a fiducial when the environment for calibration is unknown. Extra work to exclude false detection would also complicate the calibration procedure.

\begin{figure*}
    \centering
    \includegraphics[width=\linewidth]{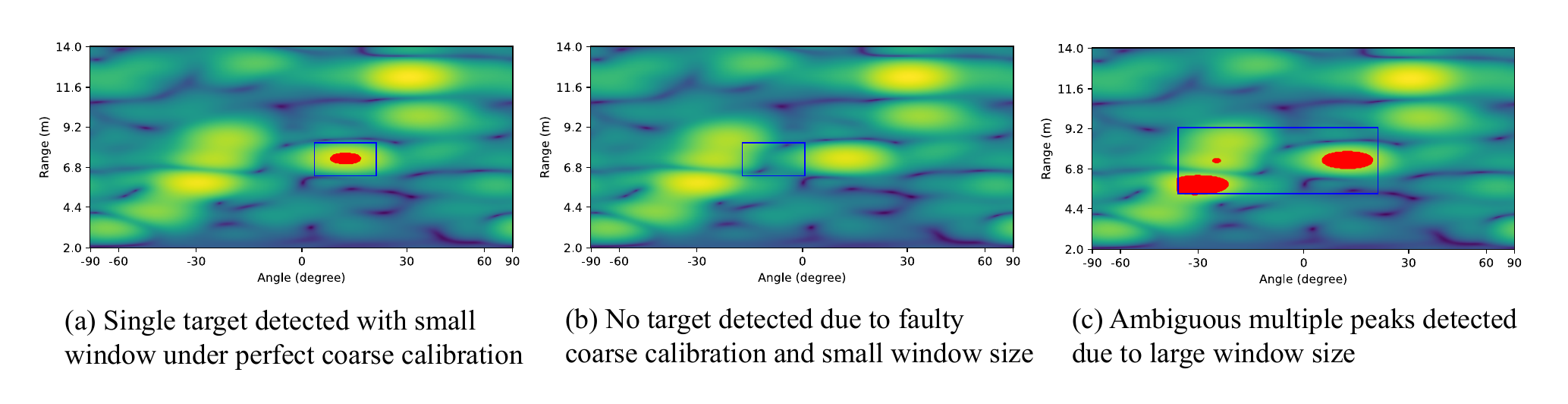}
    \caption{Potential scenarios when initial coarse calibration is used in the process of calibration with corner reflector}
    \label{fig:window_tradeoff}
\end{figure*}

\subsection{Corner reflector coarse calibration}
The corner reflector-based calibration needs a coarse calibration step~\cite{pervsic2017extrinsic} that estimates the rotation and translation of sensors to create a window in the Range-Angle FFT profile. In this evaluation, we will show how this requirement can cause problems for calibration.
 
\textbf{Effect of window size on peak detection:} Accurately measuring the relative position and angular difference between two sensors is challenging, especially with sensor rotation. A large window size have better tolerance to coarse calibration errors, but it enhances the risk of capturing multiple peaks in cluttered environments. Figure \ref{fig:window_tradeoff} depicts the trade-off between reducing manual measurement errors and minimizing false detection. \ref{fig:window_tradeoff}a shows the ideal case where an appropriately sized window captures the peak accurately. However, if the coarse calibration error exceeds the tolerance, the target will fall outside of the window, as shown in Figure \ref{fig:window_tradeoff}b. Directly widening the window may result in capturing additional peaks from surrounding objects, leading to false detection when clutter sources have higher backscatter power than the target. Figure \ref{fig:window_tradeoff}c elaborates this scenario.

\begin{figure}
\centering
\includegraphics[width=\linewidth]{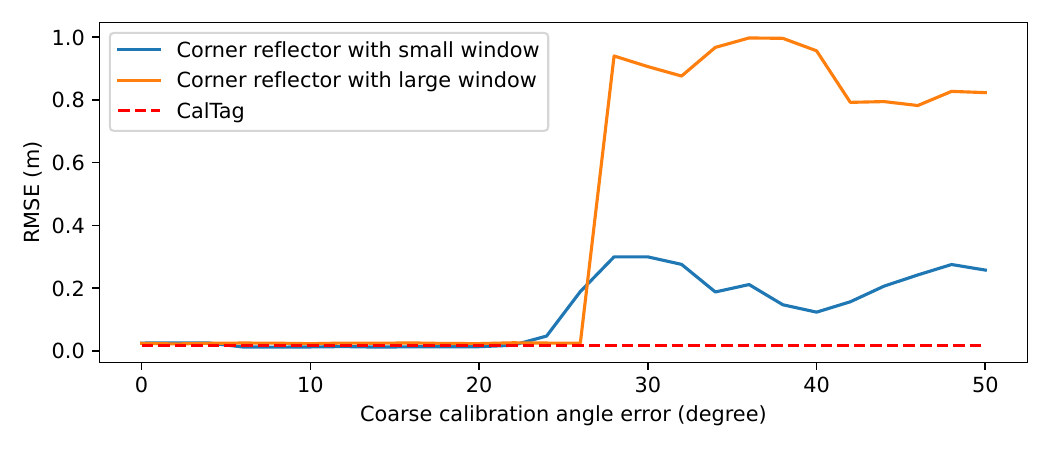}
\caption{Relationship between calibration RMSE and coarse calibration angle error under two different window sizes. The small window size is $10^\circ$ by $20cm$. The large window size is $60^\circ$ by $2m$.}
\label{fig:varying_window}
\end{figure}

\textbf{Subsequent effect on calibration error:} To better evaluate the uncertainty in the selection of window size, we conducted a calibration experiment at a medium clutter level, as depicted in \ref{fig:clutter_comparison}b, using a corner reflector and two different coarse calibration window sizes. Figure \ref{fig:varying_window} shows the calibration errors against different manual angle measurement errors. The small window is $10^\circ$ in angle by $20cm$ in range. The large window is $60^\circ$ in angle by $2m$ in range. We intentionally add angle errors to the coarse calibration matrix to simulate manual measurement flaws. The RMSE begins to increase when the angle error rises beyond $20^\circ$. The small window size results in an earlier start of increment, confirming the lower tolerance of a small window to angle errors than a large window. However, using a large window suffers from steeper increase in RMSE when false detection occurs. Even under the same environment, changes in the radar's position and direction alter the angle and the range of clutters, complicating window size selection.

\begin{figure}
\centering
\begin{subfigure}{\columnwidth}
    \centering
    \includegraphics[width=\columnwidth]{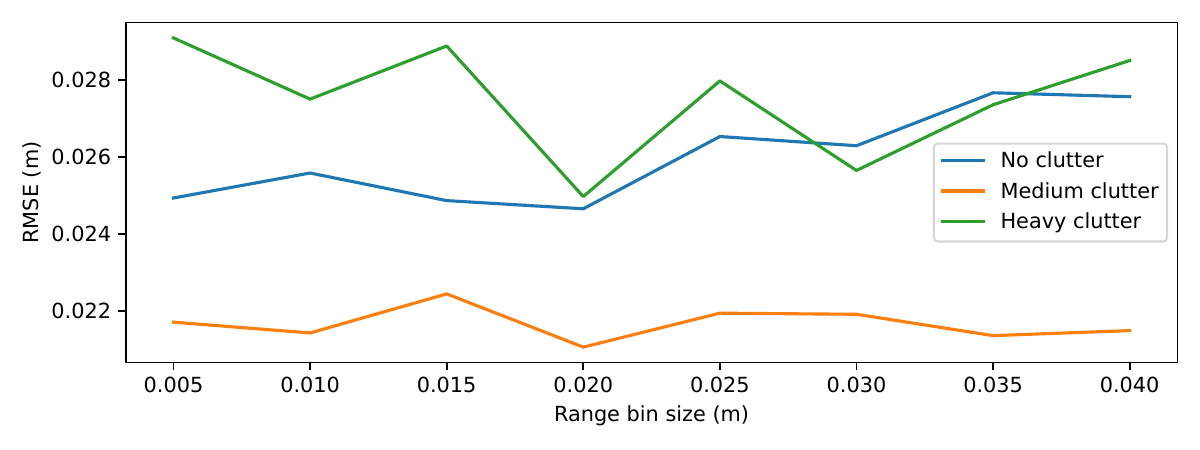}
    \subcaption{RMSE with varying range resolution, angle resolution = $0.001^\circ$}
    \label{fig:range_res}
\end{subfigure}
\begin{subfigure}{\columnwidth}
    \centering
    \includegraphics[width=\columnwidth]{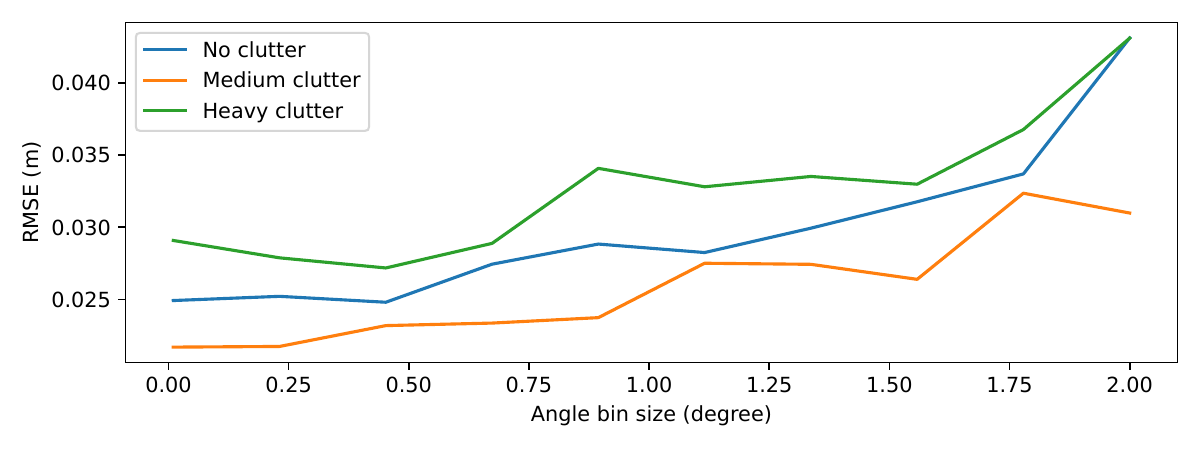}
    \subcaption{RMSE with varying angle resolution, range resolution = $0.005m$}
    \label{fig:angle_res}
\end{subfigure}
\caption{Calibration accuracy of testing set under different range and angle MUSIC resolution}
\end{figure}

\subsection{\name Detection micro-benchmarks}

\textbf{Role of precision:} The precision of fiducial detection heavily influences the calibration quality. While shrinking bin width improves calibration accuracy, it requires more computational resources. In this section, we evaluate the \name detection quality by changing the signal processing resolution and find the optimal MUSIC bin size. Figure \ref{fig:range_res} and \ref{fig:angle_res} show the average calibration accuracy across different clutter levels under various MUSIC resolutions. RMSE is kept under 0.03m when the range bin width is below 0.04m. There is no visible trend when the range bin width increases. Angle resolution has a greater impact on calibration accuracy. The calibration error increases steadily with the angle bin width by roughly 0.01m from $0.001^\circ$ to $2^\circ$. The optimal range resolution is 0.02m, and the optimal angle resolution is 0.25$^\circ$.

\textbf{Performance under large rotations:} Two sensors strictly facing the same direction is rare in a real-world deployment. To test \name's robustness against radar rotation, we collected data with the radar facing angles from $0^\circ$ to $30^\circ$ in $10^\circ$ steps relative to the LiDAR. Table \ref{table:tag_calib_radar_rotate} shows that the calibration error remains consistent across all angles. The training RMSE is around $2\pm0.5$cm. The testing RMSE is around $3\pm1$cm. The results demonstrate \name's reliability in autonomous calibration.

\begin{table}[htbp]
\footnotesize
\caption{\name calibration error under different Radar rotation angle}
\label{table:tag_calib_radar_rotate}
\centering
\begin{adjustbox}{max width=\columnwidth}
\begin{tabular}{|c|c|c|}
\hline
Rotation angle (degree) & train RMSE (m) & test RMSE (m)\\
\hline
10 & $0.0207\pm0.0053$ & $0.0276\pm0.0115$ \\
\hline
10 & $0.0203\pm0.0053$ & $0.0288\pm0.0111$ \\
\hline
20 & $0.0216\pm0.0048$ & $0.0268\pm0.0099$ \\
\hline
30 & $0.0201\pm0.0055$ & $0.0288\pm0.0177$ \\
\hline
\end{tabular}
\end{adjustbox}
\end{table}
\section{Conclusion}

Apart from traditional methods, \name and its detection method changes the backscatter signal to enhance its sensitivity to the radar. Experiments across three levels of clutter demonstrate \name's reliability. On the other hand, tests with the corner reflector under different coarse calibration angle errors show the uncertainties from potential human measurement errors, which is perfectly solved by \name. Future work can explore real-time and autonomous calibration using \name, integrate it with other sensor fiducials for more effective multi-sensor calibration, and extend the 2D calibration to 3D calibration with 6 Degrees of Freedom using radars that provide elevation data.

\bibliographystyle{plain}
\bibliography{references}

\end{document}